\documentclass[10pt,twocolumn,letterpaper]{article}

\usepackage{iccv}
\usepackage{times}
\usepackage{epsfig}
\usepackage{graphicx}
\usepackage{amsmath}
\usepackage{amssymb}
\usepackage{multirow}
\usepackage{tabularx}
\usepackage{booktabs}

\usepackage{setspace}

\usepackage[pagebackref=true,breaklinks=true,letterpaper=true,colorlinks,bookmarks=false]{hyperref}

\iccvfinalcopy % *** Uncomment this line for the final submission

\def\iccvPaperID{3382} % *** Enter the ICCV Paper ID here

\ificcvfinal\fi

\begin{document}

%%%%%%%%% TITLE
\title{BMN: Boundary-Matching Network for Temporal Action Proposal Generation}

\author{Tianwei Lin, Xiao Liu, Xin Li, Errui Ding, Shilei Wen\\
Department of Computer Vision Technology (VIS), Baidu Inc.\\
{\tt\small \{lintianwei01,liuxiao12,lixin41,dingerrui,wenshilei\}@baidu.com}
}

\maketitle
\ificcvfinal\thispagestyle{empty}\fi

\begin{abstract}

Temporal action proposal generation is an challenging and promising task which aims to locate temporal regions in real-world videos where action or event may occur.
Current bottom-up proposal generation methods can generate proposals with precise boundary, but cannot efficiently generate adequately reliable confidence scores for retrieving proposals.
To address these difficulties, we introduce the \textbf{Boundary-Matching (BM) mechanism} to evaluate confidence scores of densely distributed proposals, which denote a proposal as a matching pair of starting and ending boundaries and combine all densely distributed BM pairs into the BM confidence map.
Based on BM mechanism, we propose an effective, efficient and end-to-end   proposal generation method, named \textbf{Boundary-Matching Network (BMN)}, which generates proposals with precise temporal boundaries as well as reliable confidence scores simultaneously.
The two-branches of BMN are jointly trained in an unified framework.
We conduct experiments on two challenging datasets: THUMOS-14 and ActivityNet-1.3, where BMN shows significant performance improvement with remarkable efficiency and generalizability. Further, combining with existing action classifier, BMN can achieve state-of-the-art temporal action detection performance.

\end{abstract}

%%%%%%%%% BODY TEXT
\section{Introduction}

With the number of videos in Internet growing rapidly, video content analysis methods have attracted widespread attention from both academia and industry.
Temporal action detection is an important task in video content analysis area, which aims to locate action instances in untrimmed long videos with both action categories and temporal boundaries.
Akin to object detection, temporal action detection method can be divided into two stages: temporal action proposal generation and action classification.
Although convincing classification accuracy can be achieved by action recognition methods, the detection performance is still low in mainstream benchmarks \cite{jiang2014thumos,caba2015activitynet}. Therefore, many recent methods work on improving the quality of temporal action proposals.  Besides being used in temporal action detection task, temporal proposal generation methods also have wide applications in many areas such as video recommendation, video highlight detection and smart surveillance.

\begin{figure}[t]
\setlength{\abovecaptionskip}{-0.1cm} %缩小caption和图像之间的距离
\setlength{\belowcaptionskip}{-0.3cm} %缩小caption和下方文字的距离
\begin{center}
\begin{minipage}[b]{1.0\linewidth}
  \centering
  \centerline{\includegraphics[width=8.5cm]{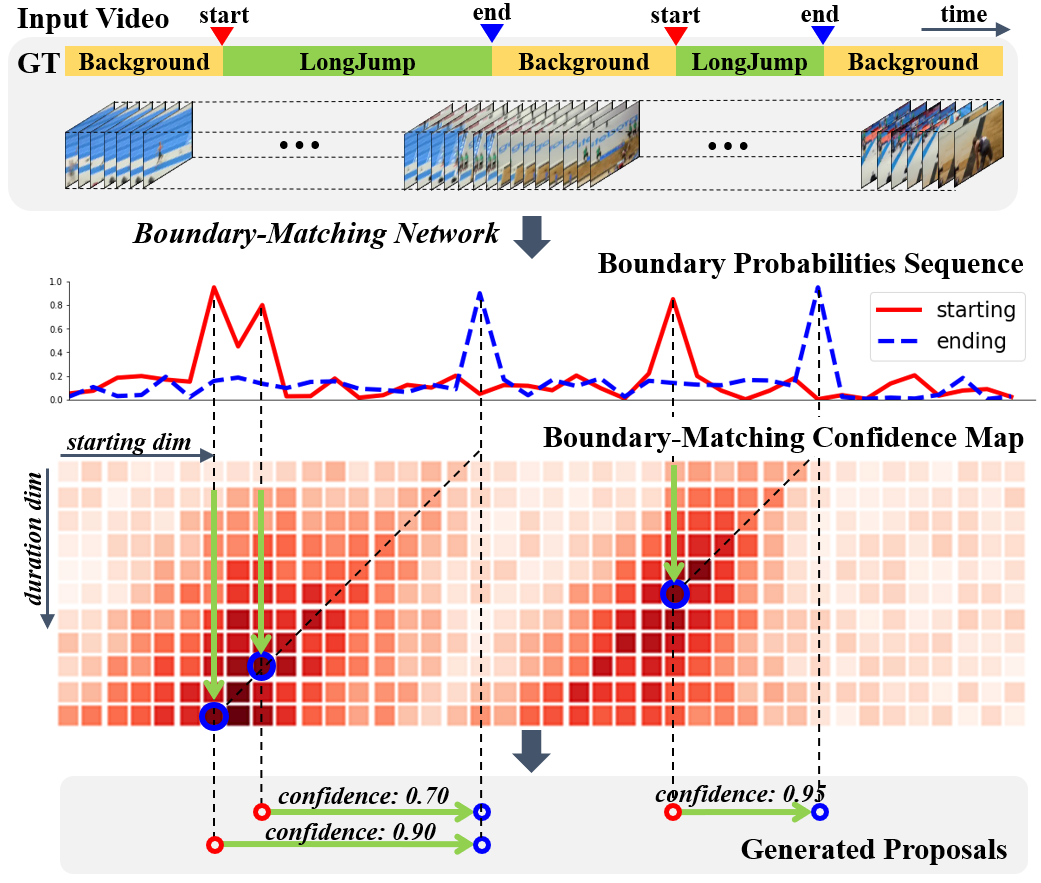}}
  \medskip
\end{minipage}
\end{center}
\vspace{-0.5cm}
   \caption{Overview of our method. Given an untrimmed video, BMN can simultaneously generate (1) boundary probabilities sequence to construct proposals and (2) Boundary-Matching confidence map to densely evaluate confidence of all proposals.}
\label{fig:overview}
\vspace{-0.3cm}
\end{figure}

To achieve high proposal quality, a proposal generation method should (1) generate temporal proposals with flexible duration and precise boundaries to cover ground-truth action instances precisely and exhaustively; (2) generate reliable confidence scores so that proposals can be retrieved properly.
Most existing proposal generation methods \cite{sst_buch_cvpr17,fast_temporal_activity_cvpr16,escorcia2016daps,shou2016action} adopted a \textbf{``top-down''} fashion to generate proposals with multi-scale temporal sliding windows in regular interval, and then evaluate confidence scores of proposals respectively or simultaneously.
The main drawback of these methods is that generated proposals are usually not temporally precise or not flexible enough to cover  ground-truth action instances of varies duration.
Recently, Boundary-Sensitive Network (BSN) \cite{lin2018bsn} adopted a \textbf{``bottom-up''} fashion to generate proposals in two stages:  (1) locate temporal boundaries and combine boundaries as proposals and (2) evaluate confidence score of each proposal using constructed proposal feature. 
By exploiting local clues, BSN can generate proposals with  more precise boundaries and more  flexible duration than existing top-down methods.
However, BSN has three main drawbacks: (1) proposal feature construction and confidence evaluation procedures are conducted to each proposal respectively, leading to inefficiency;
(2) the proposal feature constructed in BSN is too simple to capture enough temporal context;
(3) BSN is multiple-stage but not an unified framework.

Can we evaluate confidence for all proposals simultaneously with rich context?
Top-down methods \cite{ssad,sstad} can achieve this easily with anchor mechanism, where proposals are pre-defined as non-continuous distributed anchors.
However, since the boundary and duration of proposals are much more flexible, anchor mechanism is not suitable for bottom-up methods such as BSN. 
To address these difficulties, we propose the \textbf{Boundary-Matching (BM) mechanism} for  confidence evaluation of densely distributed proposals. In BM mechanism, a proposal is denoted as a matching pair of its starting and ending boundaries, and then all BM pairs are combined  as a two dimensional BM confidence map to represent densely distributed proposals with continuous starting boundaries and temporal duration.
%where two dimensions indicate starting time  and temporal duration  separately.
%
Thus, we can generate confidence scores for all proposals simultaneously via the BM confidence map.
A BM layer is proposed to generate BM feature map from temporal feature sequence, and the BM confidence map can be obtained from the BM feature map using a series of conv-layers. BM feature map contains rich feature and temporal context for each proposal, and gives the potential for exploiting context of adjacent proposals.

In summary, our work has three main contributions:

\begin{enumerate}
\setlength{\itemsep}{0pt}
\setlength{\parsep}{0pt}
\setlength{\parskip}{0pt}

\item We introduce the \emph{Boundary-Matching mechanism} for evaluating confidence scores  of densely distributed proposals, which can be easily embedded in network.

\item We propose an efficient, effective and end-to-end temporal action proposal generation method \emph{Boundary-Matching Network} (BMN). Temporal boundary probability sequence and BM confidence map are generated simultaneously in two branches of BMN, which are trained jointly as an unified framework.

\item Extensive experiments show that BMN can achieve significantly better proposal generation performance than other state-of-the-art methods, with remarkable efficiency, great generalizability and great performance on temporal action detection task.
 
\end{enumerate}

\section{Related Work}

\noindent
\textbf{Action Recognition.}
Action recognition is a fundamental and important task of video understanding area.
Hand-crafted features such as HOG, HOF and MBH are widely used in earlier works, such as improved Dense Trajectory (iDT) \cite{dtf,wang2013action}.
Recently, deep learning models have achieved significantly performance promotion in action recognition task. 
The mainstream networks fall into two categories: two-stream networks \cite{feichtenhofer2016convolutional,simonyan2014two,wang2015towards} exploit appearance and motion clues from RGB image and stacked optical flow separately; 3D networks \cite{tran2015learning,qiu2017learning}  exploit appearance and motion clues directly from raw video volume.
In our work, by convention, we adopt action recognition models to extract visual feature sequence of untrimmed video.

\noindent
\textbf{Correlation Matching.}
Correlation matching algorithms are widely used in many computer vision tasks, such as image registration, action recognition and stereo matching.
%Stereo matching is a fundamental computer vision task, which aims to find corresponding pixels from stereo images.
%
Specifically, stereo matching aims to find corresponding pixels from stereo images. For each pixel in left image of a rectified image pair, the stereo matching method need to find corresponding pixel in right image along horizontal direction, or we can say finding right pixel with minimum cost.
Thus, the cost minimization of all left pixels can be denoted as a cost volume, which denotes each left-right pixel pair as a point in volume.
Based on cost volume, many recent works \cite{song2018edgestereo,mayer2016large,liang2017learning} achieve end-to-end network via generating cost volume directly from combining two feature maps, using correlation layer \cite{mayer2016large} or feature concatenation \cite{chang2018pyramid}.
Inspired by cost volume, our proposed BM confidence map contains pairs of temporal starting and ending boundaries as proposals, thus can directly generate confidence scores for all proposals using convolutional layers.
We propose BM layer to efficiently generate BM feature map via sampling feature among starting and ending boundaries of each proposal simultaneously.

\noindent
\textbf{Temporal Action Proposal Generation.}
As aforementioned, the goal of temporal action detection task is to detect action instances in untrimmed videos with temporal boundaries and action categories, which can be divided into temporal proposal generation and action classification stages.
These two stages are taken apart in most detection methods \cite{shou2016action,singh2016untrimmed,zhao2017temporal}, and are taken together as single model in some methods \cite{ssad,sstad}.
For proposal generation task, most previous works \cite{sst_buch_cvpr17,fast_temporal_activity_cvpr16, escorcia2016daps, gao2017turn,shou2016action}  adopt \emph{top-down} fashion to generate proposals with pre-defined duration and interval, where the main drawback is the lack of boundary precision and duration flexibility.
There are also some methods \cite{zhao2017temporal,lin2018bsn} adopt \emph{bottom-up} fashion. TAG \cite{zhao2017temporal} generates proposals using temporal watershed algorithm, but lack confidence scores for retrieving. Recently, BSN \cite{lin2018bsn} generates proposals via locally locating temporal boundaries and globally evaluating confidence scores, and achieves significant performance promotion over previous proposal generation methods.
In this work, we propose the Boundary-Matching mechanism for proposal confidence evaluation, which can largely simplify the pipeline of BSN and bring significant promotion in both efficiency and effectiveness.

\section{Our Approach}

\subsection{Problem Formulation}

We can denote an untrimmed video $X$ as frame sequence $X=\left \{ x_n \right \}_{n=1}^{l_v}$ with $l_v$  frames, where $x_n$ is the $n$-th RGB frame of video $X$.
The temporal annotation set of $X$ is composed by
a set of temporal action instances as  $\Psi_g  = \left \{ \varphi  _n=\left (t_{s,n},t_{e,n}   \right ) \right \}_{n=1}^{N_g}$, where $N_g$ is the amount of ground-truth action instances,  $t_{s,n}$ is the starting time of action instance $\varphi_n$ and $t_{e,n} $ is the ending time.
Unlike temporal action detection task, categories of action instances are not taken into account in proposal generation task.
During inference, proposal generation method should generate proposals $\Psi_p$ which cover $\Psi_g$ precisely and exhaustively.

\subsection{Feature Encoding.}

Following recent proposal generation methods \cite{sst_buch_cvpr17, escorcia2016daps, gao2017turn,lin2018bsn}, we construct BMN model upon visual feature sequence extracted from raw video. 
In this work, we adopt two-stream network \cite{simonyan2014two} for feature encoding since it achieves great action recognition precision and is widely used in many video analysis methods \cite{gao2017cascaded,ssad,zhao2017temporal}.
Concatenating the output scores of top fc-layer in two-stream network, we can get encoded visual feature $f_{t_n} \in R^{C}$ around frame $x_{t_n}$, where $C$ is the dimension of feature.
Therefore, given an untrimmed video $X$ of length $l_v$, we can extract a visual feature sequence $F=\left \{ f_{t_n} \right \}_{n=1}^{l_f} \in R^{C\times l_f}$ with length $l_f$.
To reduce the computation cost, we extract feature in a regular frame interval $\sigma$, thus $l_f=l_v/\sigma$.

\begin{figure}[t]
\begin{center}
\begin{minipage}[b]{1.0\linewidth}
  \centering
  \centerline{\includegraphics[width=8.5cm]{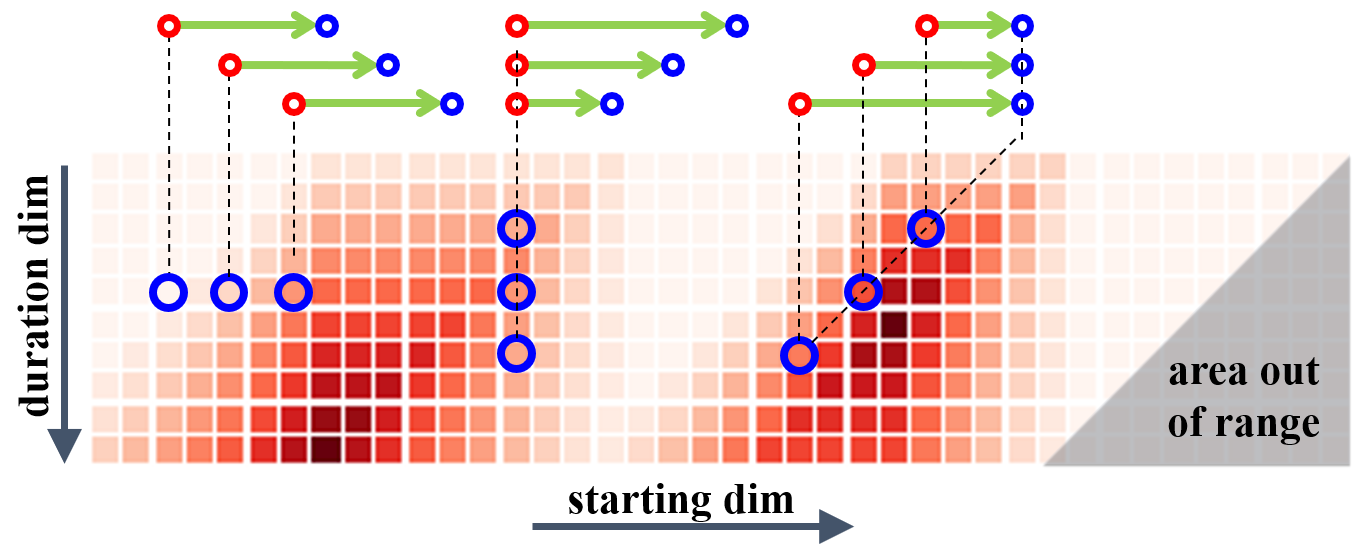}}
  \medskip
\end{minipage}
\end{center}
   \caption{Illustration of BM confidence map. Proposals in the same row have the same temporal duration, and proposals in the same column have the same starting time. The ending boundaries of proposals at right-bottom corner exceed the range of  video, thus these proposals are not considered during training and inference.}
\label{fig:bm_mechanism}
\end{figure}

\subsection{Boundary-Matching Mechanism}

In this section, we introduce the Boundary-Matching (BM) mechanism to generate confidence scores for densely distributed proposals. 
First we denote a temporal proposal $\varphi$ as a matching pair of its starting boundary $t_s$ and  ending boundary $t_e$. 
Then, as shown in Fig \ref{fig:bm_mechanism}, the goal of BM mechanism is to generate the two dimensional BM confidence map $M_C$, which is constructed by BM pairs with different starting boundary and temporal duration.
In BM confidence map, the value of point $M_C (i,j)$ is denoted as the confidence score of  proposal $\varphi_{i,j}$ with starting boundary $t_s = t_j$, duration $d = t_i$ and ending boundary  $t_e = t_j + t_i$. 
Thus, we can generate confidence scores for densely distributed proposals via generating BM confidence map.

\begin{figure}[t]
\begin{center}
\begin{minipage}[b]{1.0\linewidth}
  \centering
  \centerline{\includegraphics[width=8.5cm]{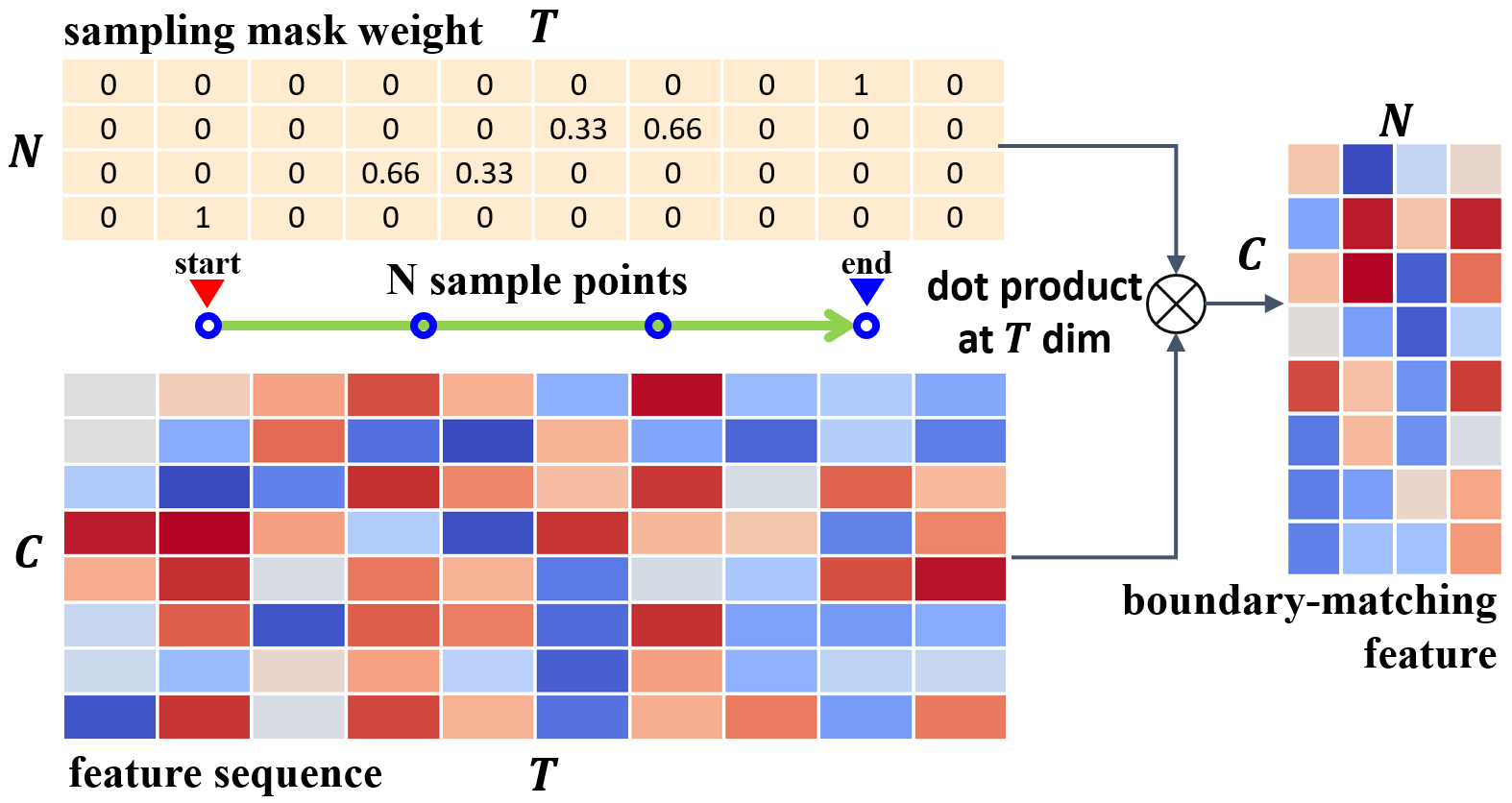}}
  \medskip
\end{minipage}
\end{center}
   \caption{Illustration of BM layer. For each proposal, we conduct dot product at $T$ dimension between sampling weight and temporal feature sequence, to generate BM feature of shape $C\times N$.}.
   \label{fig:bm_layer}
\end{figure}

\noindent
\textbf{Boundary-Matching Layer.} 
How can we generate two dimensional BM confidence map from temporal feature sequence?
In BM mechanism, we introduce the BM layer to generate BM feature map $M_F \in R^{C \times N \times D \times T}$ from temporal feature sequence $S_F \in R^{C \times T}$, and then use $M_F$ to generate BM confidence map $M_C \in R^{D\times T}$ with a series of convolutional layers, where $D$ are pre-defined maximum proposal duration.
The goal of BM layer is to uniformly sample $N$ points in $S_F$ between starting boundary $t_s$ and ending boundary $t_e$ of each proposal $\varphi_{i,j}$, and get proposal feature $m^f_{i,j} \in R^{C\times N}$ with rich context.
And we can generate BM feature map $M_F$ via conducting this sampling procedure for all proposals simultaneously.

There are two difficulties to achieve this feature sampling procedure: (1) how to sample feature in non-integer point and (2) how to sample feature for all proposals simultaneously.
As shown in Fig \ref{fig:bm_layer}, we achieve this via dot product between temporal feature sequence $S_F \in R^{C\times T}$ and sampling mask weight  $W  \in R^{N\times T \times D \times T}$ in temporal dimension.
In detail, \textbf{first}, for each proposal $\varphi_{i,j}$, we construct  weight term $w_{i,j} \in R^{N\times T}$ via uniformly sampling $N$ points between expanded temporal region $[ t_s - 0.25d, t_e + 0.25d]$. 
For a non-integer sampling point $t_n$, we define its corresponding sampling mask $w_{i,j,n} \in R^T$ as

\begin{equation}
w_{i,j,n}[t]=
\begin{cases}
1 - dec(t_n) & \text{ if } t= floor(t_n)\\ 
dec(t_n) & \text{ if } t= floor(t_n) +1,\\ 
0 & \text{ if } t= others
\end{cases}
\end{equation}
where $dec$ and $floor$ is decimal and integer fraction functions separately.
Thus, for proposal $\varphi_{i,j}$, we can get weight term $w_{i,j} \in R^{N\times T}$.
\textbf{Second}, we conduct dot product in temporal dimension between $S_F$ and $w_{i,j}$
 
\begin{equation}
 m^f_{i,j}[c,n] = \sum_{t=1}^T S_f[c,t] \cdot  w_{i,j}[n,t].
\end{equation}

Via expanding $w_{i,j} \in R^{N\times T}$ to $W  \in R^{N\times T \times D \times T}$ for all proposals in BM confidence map, we can generate BM feature map $M_F \in R^{C \times N \times D \times T}$ using dot product. 
Since the sampling mask weight $W$ is the same for different videos and can be pre-generated, the inference speed of BM layer is very fast.
BM feature map contains rich feature and temporal context for each proposal, and gives the potential for exploiting context of adjacent proposals.
%
%The procedure of generating $M_C$ using $M_F$ is introduced in section 3.4.

\begin{figure}[t]
\begin{center}
\begin{minipage}[b]{1.0\linewidth}
  \centering
  \centerline{\includegraphics[width=8.5cm]{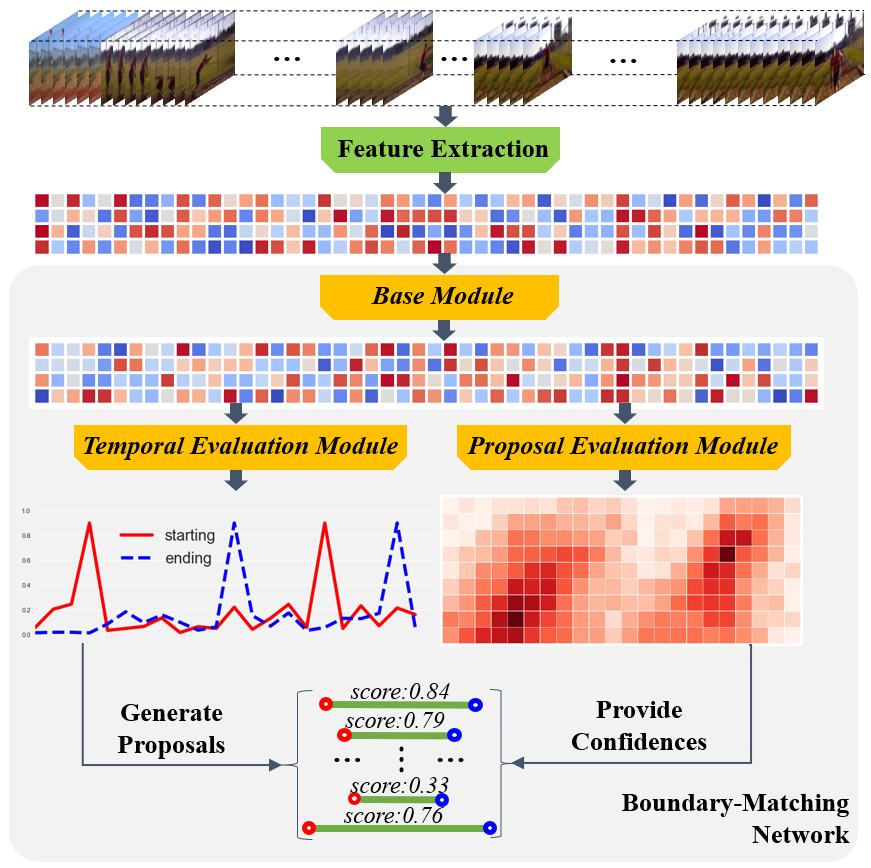}}
  \medskip
\end{minipage}
\end{center}
   \caption{The framework of Boundary-Matching Network. After feature extraction, we use BMN to simultaneously generate temporal boundary probability sequence and BM confidence map, and then construct proposals based on boundary probabilities and get corresponding confidence score from BM confidence map.}
\label{fig:bmn_framework}
\end{figure}

\noindent
\textbf{Boundary-Matching Label.} 
During training, we  denote the BM label map as $G_C \in R^{D\times T}$ with the same shape of BM confidence map $M_C$, where $g^c_{i,j} \in [0,1]$ represents the maximum $IoU$ between proposal $\varphi_{i,j}$ and all ground-truth action instances.
Generally, in BM mechanism, we use BM layer to efficiently generate BM feature map $M_F $ from temporal feature sequence $S_F $, and then use a series of convolutional layers to generate BM confidence map $M_C $, which is trained under supervision of BM label map $G_C$.

\subsection{Boundary-Matching Network}

Different with the multiple-stage framework of  BSN \cite{lin2018bsn}, BMN generates local boundary probabilities sequence and global proposal confidence map simultaneously, while the whole model is trained in an unified framework.
As demonstrated in Fig \ref{fig:bmn_framework}, BMN model contains three modules: \emph{Base Module } handles the input feature sequence, and outputs feature sequence shared by the following two modules; \emph{Temporal Evaluation Module } evaluates starting and ending probabilities of each location in video to generate boundary probability sequences; 
\emph{Proposal Evaluation Module } contains the BM layer to transfer feature sequence to  BM feature map, and contains a series of 3D and 2D convolutional layers to generate BM confidence map.

\begin{table}[tbp]
\centering
\caption{The detailed architecture of BMN, where the output feature sequence of base module is shared by temporal evaluation and proposal evaluation modules. $T$ and $D$ are length of input feature sequence and maximum proposal duration separately.   }
\small
\begin{tabular}{p{1.28cm}|p{0.6cm}p{0.6cm}p{0.3cm}p{1.0cm}|p{2.0cm}}
%\multicolumn{6}{c}{ {\bf ActivityNet-1.3}, mAP@$tIoU$}  \\
\hline
 layer & kernel & stride & dim & act & output size\\
\hline
\multicolumn{6}{c}{ Base Module} \\
\hline
 $conv1d_1$ & 3 &  1 &  256 & $relu$ & 256$\times T$\\
\hline
 $conv1d_2$ & 3 &  1 &  128 & $relu$ & 128$\times T$\\
\hline
\multicolumn{6}{c}{ Temporal Evaluation Module} \\
\hline
 $conv1d_3$ & 3 &  1 &  256 & $relu$ & 256$\times T$\\
\hline
 $conv1d_4$ &  3 &  1 &  2 & $sigmoid$  & 2$\times T$\\
\hline
\multicolumn{6}{c}{ Proposal Evaluation Module}\\
\hline
 BM layer & \multicolumn{4}{c|}{ N - 32}  & 128$\times$32$\times D$$\times T$\\
\hline
 $conv3d_1$ & 32,1,1 & 32,0,0 & 512 & $relu$  & 512$\times$1$\times D$$\times T$\\
\hline
 squeeze & & & & & 512$\times D$$\times T$\\
\hline
 $conv2d_1$ & 1,1 & 0,0 & 128 & $relu$   & 128$\times D$$\times T$\\
\hline
 $conv2d_2$ & 3,3 & 1,1 & 128 & $relu$   & 128$\times D$$\times T$\\
\hline 
 $conv2d_3$ & 1,1 & 0,0 & 2 & $sigmoid$  &  2$\times D$$\times T$\\
\hline
\end{tabular}
\label{table:bmn_layer}
\end{table}

\noindent
\textbf{Base Module.}
The goal of the base module is to handle the input feature sequence, expand the receptive field and serve as backbone of network, to provide a shared feature sequence for TEM and PEM.
Since untrimmed videos have uncertain temporal length, we adopt a long observation window with length $l_{\omega}$ to truncate the untrimmed feature sequence with length $l_f$.
We denote an observation window as  $\omega =\left \{ t_{\omega,s}, t_{\omega,e}, \Psi_{\omega}, F_{\omega}  \right \}$, where $t_{\omega,s}$ and $t_{\omega,e}$ are the starting and ending time of $\omega$ separately, $\Psi_{\omega}$ and $F_{\omega}$   are annotations and feature sequence  within the window separately.
The window length $l_{\omega} = t_{\omega,e} - t_{\omega,s}$ is set depending on the dataset.
The details of base module is shown in Table \ref{table:bmn_layer}, including two temporal convolutional layers.

\noindent
\textbf{Temporal Evaluation Module (TEM).}
The goal of TEM is to evaluate the starting and ending probabilities for all temporal locations in untrimmed video. These boundary probability sequences are used for generating proposals during post processing.
The details of TEM are shown in Table \ref{table:bmn_layer}, where $conv1d_4$ layer with two sigmoid activated filters output starting probability sequence $P_{S,\omega}=\left \{ p^s_{t_n} \right \}_{n=1}^{l_{\omega}}$ and ending probability sequence $P_{E,\omega}=\left \{ p^e_{t_n} \right \}_{n=1}^{l_{\omega}}$ separately for an observation window ${\omega}$.

\noindent
\textbf{Proposal Evaluation Module (PEM).}
The goal of PEM is to generate Boundary-Matching (BM) confidence map, which contains  confidence scores for densely distributed proposals. To achieve this, PEM contains BM layer and a series of 3d and 2d convolutional layers.

As introduced in Sec. 3.3, BM layer transfers temporal feature sequence $S$ to BM feature map $M_F$ via matrix dot product between  $S$ and sampling mask weight $W$ in temporal dimension.
In BM layer, the number of  sample points $N$ is set to 32, and the maximum proposal duration $D$ is set depending on dataset.
After generating BM feature map $M_F$, first we conduct $conv3d_1$ layer in sample dimension to reduce dimension length from $N$ to $1$, and increase hidden units from $128$ to $512$. 
Then, we conduct $conv2d_1$ layer with $1\times1$ kernel to reduce the hidden units, and $conv2d_2$ layer with $3\times3$ kernel to capture context of adjacent proposals. Finally, we generate two types of BM confidence map $M_{CC}, M_{CR} \in R^{D\times T}$ with \emph{sigmoid} activation, where $M_{CC}$ and $M_{CR}$ are trained using binary classification and regression loss function separately.

\subsection{Training of BMN}

In BMN, TEM learns local boundary context and PEM pattern global proposal  context.
To jointly learn local pattern and global pattern, an unified multi-task framework is exploited for optimization. The training details of BMN are introduced in this section.

\noindent
\textbf{Training Data Construction.}
Given an untrimmed video $X$, we can extract feature sequence $F$ with length $l_f$.
Then, we use observation windows with length $l_{\omega}$ to truncate feature sequence with $50\%$ overlap, where windows containing at least one ground-truth action instance are kept for training.
Thus, a training set  $\Omega= \left \{ \omega_n \right \}_{n=1}^{N_{\omega }}$ is constructed with $N_{\omega}$ observation windows.

\noindent
\textbf{Label Assignment.}
%To train BMN, first we need generate boundary label sequence $G_S, G_E \in R^T$ and BM confidence label map $G_M$.
For TEM, we need to generate temporal boundary label sequence $G_S, G_E \in R^T$.
Following BSN\cite{lin2018bsn}, for a ground-truth action instance $\varphi_g=( t_{s},t_{e} )$ with duration $d_g=t_e-t_s$ in annotation set $\Psi_{\omega}$, we denote its starting and ending regions as $r_S=[ t_s-d_g/10,t_s+d_g/10 ]$ and $r_E= [ t_e-d_g/10,t_e+d_g/10 ]$ separately. 
Then, for a temporal location $t_n$ within $F_{\omega}$, we denote its local region as $r_{t_n}=[ t_n-d_f/2,t_n+d_f/2 ]$, where $d_f=t_{n}-t_{n-1}$ is the temporal interval between two locations. 
Then we calculate overlap ratio $IoR$ of $r_{t_n}$ with $r_S$ and $r_E$ separately, and denote maximum $IoR$ as $g^s_{t_n}$ and $g^e_{t_n}$ separately, where $IoR$ is defined as the overlap
ratio with groundtruth proportional to the duration of this region.
Thus we can generate $G_{S,\omega} = \left \{ g^s_{t_n} \right \}_{n=1}^{l_{\omega }} $
 and $G_{E,\omega} = \left \{ g^e_{t_n} \right \}_{n=1}^{l_{\omega }} $ as label of TEM.

For PEM, we need to generate BM  label map $G_C \in R^{D \times T}$.
For a proposal $\varphi_{i,j} =(t_s = t_j, t_e = t_j + t_i)$, we calculate its Intersection-over-Union ($IoU$) with all $\varphi_g$ in $\Psi_{\omega}$, and denote the maximum $IoU$ as $g^c_{i,j}$.
Thus we can generate $G_C = \left \{ g^c_{i,j} \right \}_{i,j=1}^{D,l_{\omega }} $ as label of PEM.

\noindent
\textbf{Loss of TEM.}
With generated boundary probability sequence $P_{S,\omega}$, $P_{E,\omega}$ and boundary label sequence $G_{S,\omega}$, $G_{E,\omega}$, we can construct the  loss function of TEM  as the sum of staring and ending losses

\begin{equation}
L_{TEM}= L_{bl}(P_{S},G_{S})+   L_{bl}(P_{E},G_{E}).
\end{equation}

Following BSN\cite{lin2018bsn}, we adopt weighted binary logistic regression loss function $L_{bl}$ for both starting and ending losses, where $L_{bl}(P,G)$ is denoted as:

\begin{small}
\begin{equation}
\frac{1}{l_\omega}\sum_{i=1}^{l_\omega} \left (\alpha^{+}\cdot b_i\cdot  log(p_i)+\alpha^{-} \cdot (1-b_i) \cdot log(1-p_i) \right ),
\end{equation}
\end{small}
where $b_i=sign(g_i-\theta)$ is a two-value function used to convert  $g_i$ from $[ 0,1 ]$ to $\left \{0,1  \right \}$ based on  overlap threshold $\theta = 0.5$.  
Denoting $l^+=\sum b_i$ and $l^-=l_\omega -l^+ $, the weighted terms are $\alpha^+=\frac{l_w}{l^+}$ and $\alpha^-=\frac{l_w}{l^-}$.

%$\varphi=(t_s,t_e,p_{conf},p^s_{t_s},p^e_{t_e})$

\noindent
\textbf{Loss of PEM.}
With generated BM confidence map $M_{CC}$, $M_{CR}$ and BM label map $G_C$, we can construct the loss function of PEM, which is the sum of binary classification loss and regression loss:
% explain why we use two types of loss function

\begin{equation}
L_{PEM}=  L_{C}(M_{CC},G_C) + \lambda \cdot L_{R}(M_{CR},G_C).
\end{equation}
where we adopt $L_{bl}$ for classification loss $L_C$ and L2 loss for regression loss $L_R$, and set the weight term $\lambda=10$. 
To balance the ratio between positive and negative samples in $L_R$, we take all points with $g^c_{i,j} >0.6$ as positive and randomly sample  $g^c_{i,j} <0.2$ as negative, and ensure the ratio between positive and negative points nearly 1:1.

\noindent
\textbf{Training Objective.}
We train BMN  in the form of a multi-task loss function, including TEM loss, PEM loss and L2 regularization term:

\begin{equation}
L = L_{LEM}+\lambda_1 \cdot L_{GEM} + \lambda_2 \cdot L_2(\Theta),
\end{equation}
where weight term $\lambda_1 $ and $\lambda_2 $ are set to 1 and 0.0001 separately to ensure different modules are trained evenly.

\subsection{Inference of BMN}

During inference, we use BMN to generate boundary probability sequences $G_S$, $G_E$ and BM confidence map $M_{CC}$, $M_{CR}$. To get final results, we need to (1) generate candidate proposals using boundary probabilities, (2) fuse boundary probability and confidence score to generate final confidence score, (3) and suppress redundant proposals based on final confidence scores. 

\noindent
\textbf{Candidate Proposals Generation.}
Following BSN \cite{lin2018bsn}, we generate candidate proposals via combining  temporal locations with high boundary probabilities. 
First, to locate high starting probability locations, we record all temporal locations $t_n$ with starting $p^s_{t_n}$ (1) higher than $0.5\cdot max(p)$ or (2) being a probability peak, where $max(p^s)$ is the maximum starting probability of this video. These candidate starting locations are grouped as $B_S=\left \{ t_{s,i} \right \}_{i=1}^{N_S}$. 
We can generate ending locations set $B_E$ in the same way.

Then we match each starting location $t_s$ in $B_S$ and ending location $t_e$ in $B_E$ as a  proposal, if its duration is smaller than a pre-defined maximum duration $D$.
The generated  proposal $\varphi$ is denoted as  $\varphi =(t_{s},t_{e},p_{t_{s}}^s,p_{t_{e}}^e,p_{cc},p_{cr})$, where $p_{t_{s}}^s$, $p_{t_{e}}^e$ are starting and ending probabilities in $t_{s}$ and $t_{e}$ separately, and $p_{cc}$, $p_{cr}$ are classification confidence score and regression confidence score from $[t_e-t_s,t_s]$ point of BM confidence map $M_{CC}$ and $M_{CR}$ separately.
Thus we can get candidate proposals set $\Psi=\left \{ \varphi_i \right \}_{i=1}^{N_p}$, where $N_p$ is the number of candidate proposals.

\noindent
\textbf{Score Fusion.}
To generate more reliable confidence scores, for each proposal $\varphi$, we fuse its boundary probabilities and confidence scores by multiplication to generate the final confidence score $p_f$:

\begin{equation}
p_f = p_{t_{s}}^s \cdot p_{t_{e}}^e \cdot \sqrt{p_{cc}\cdot p_{cr}}.
\end{equation}

Thus, we can get candidate proposals set $\Psi_p=\left \{ \varphi_i = (t_s,t_e,p_f) \right \}_{i=1}^{N_p}$, where $p_f$ is used for proposals retrieving during redundant proposals suppression. %In Section xx, we explore the proposal performance with different score fusion strategy.  

\noindent
\textbf{Redundant Proposals Suppression.}
After generating candidate proposals, we need to remove redundant proposals to achieve higher recall with fewer proposals, where Non-maximum suppression (NMS) algorithm is widely used for this purpose.
In BMN, we mainly adopt Soft-NMS algorithm\cite{softNMS}, since it has proven its effectiveness in proposal generation task \cite{lin2018bsn}. Soft-NMS algorithm suppresses redundant results via decaying their confidence scores.
Soft-NMS generates suppressed final proposals set $\Psi'_p  = \left \{ \varphi  _n=(t_s,t_e,p'_f  ) \right \}_{n=1}^{N'_p}$, where $N'_p$ is the final proposals number.
During experiment, we also try normal Greedy-NMS for fair comparison.
%{{To verify the effectiveness of LGN, we also adopt normal hard-NMS during experiment in section IV.}}

\section{Experiments}

\subsection{Dataset and Setup }

\noindent
\textbf{Dataset.}
We conduct experiments on two challenging datasets: {\bf THUMOS-14} \cite{jiang2014thumos} dataset contains 413 temporal annotated untrimmed videos with 20 action categories;  {\bf ActivityNet-1.3} \cite{caba2015activitynet}  is a large-scale action understanding dataset, containing action recognition, temporal detection, proposal generation and dense captioning tasks. ActivityNet-1.3 dataset contains 19994 temporal annotated untrimmed videos with 200 action categories, which are divided into training, validation and testing sets by ratio 2:1:1.

\noindent
{\bf Implementation Details.}
For feature encoding, following previous works \cite{lin2018bsn,gao2017turn}, we adopt two-stream network \cite{xiong2016cuhk} pre-trained on training set of ActivityNet-1.3, where spatial and temporal sub-networks adopt ResNet and BN-Inception network separately.
The frame interval $\sigma $ is set  to 5 and 16 on THUMOS-14 and ActivityNet-1.3 separately.

On THUMOS-14, we set the length of observation window $l_\omega$ to 128 and the maximum duration length $D$ to 64, which can cover length of $98\%$ action instances.
On ActivityNet, following \cite{lin2018bsn,lin2017temporal}, we rescale each feature sequence to the length of the observation window $l_\omega = 100$ using linear interpolation, and the duration of corresponding annotations to range [0,1]. The maximum duration length $D$ is set to 100, which can cover length of all action instances.
%
%We implement BMN based on  Pytorch [xx].
%
To train BMN from scratch, we set learning rate to 0.001, batch size to 16 and epoch number to 10 for both datasets.

\subsection{Temporal Action Proposal Generation}

\begin{table}[tbp]
\centering
\caption{ Comparison between our method and other state-of-the-art temporal action proposal generation methods on validation set of  ActivityNet-1.3 dataset in terms of AR@AN and AUC. }
\small
%\footnotesize
\begin{tabular}{m{1.9cm}m{0.55cm}<{\centering}m{0.55cm}<{\centering}m{0.55cm}<{\centering}m{0.55cm}<{\centering}m{0.55cm}<{\centering}m{0.55cm}<{\centering}}
\toprule
Method  &  \cite{dai2017temporal}   &  \cite{ghanem2017activitynet} & \cite{lin2017temporal}  & \cite{gao2018ctap} &  \cite{lin2018bsn} & BMN \\
\hline 
AR@100 (val)	& -		& - 	& 73.01	& 73.17 & 74.16 & {\bf 75.01}  \\
AUC (val) 		& 59.58 & 63.12 & 64.40 & 65.72	& 66.17 & {\bf 67.10} \\
AUC (test) 		& 61.56	& 64.18 & 64.80 & - 	& 66.26 & {\bf 67.19}  \\
\bottomrule
\end{tabular}
\label{table:proposal_anet}
\end{table}

\begin{table}[tbp]
\centering
\caption{ Comparison between our method with state-of-the-art proposal generation methods SCNN  \cite{shou2016action}, SST \cite{sst_buch_cvpr17}, TURN \cite{gao2017turn}, TAG \cite{zhao2017temporal}, CTAP \cite{gao2018ctap}, BSN \cite{lin2018bsn} on  THUMOS-14 dataset in terms of AR@AN, where SNMS stands for Soft-NMS.} %Ablation study of input feature and NMS are also implement to confirm the effectiveness of LGN itself. 
\small
\begin{tabular}{m{1.0cm}<{\centering}m{1.6cm}m{0.55cm}<{\centering}m{0.55cm}<{\centering}m{0.55cm}<{\centering}m{0.55cm}<{\centering}m{0.6cm}<{\centering}}
\toprule
Feature & Method  		& @50 & @100  & @200 & @500 & @1000    \\
\hline 
%C3D & DAPs [6] & 23.83 &  33.96 & 57.64 & - & -  \\
C3D & SCNN-prop & 17.22 & 26.17 &  37.01 & 51.57 & 58.20   \\
C3D & SST		& 19.90 & 28.36 &  37.90 & 51.58 & 60.27   \\
C3D & TURN 		& 19.63 & 27.96 &  38.34 & 53.52 & 60.75 \\
C3D & BSN+NMS 	& 27.19 & 35.38 &  43.61 &  53.77  & 59.50   \\
C3D & BSN+SNMS 	& 29.58 & 37.38 &  45.55 &  54.67  & 59.48   \\
\hline
C3D & BMN+NMS	&  29.04 &  37.72 & 46.79 & 56.07 & {\bf 60.96}  \\
C3D & BMN+SNMS 	& {\bf 32.73} & {\bf 40.68} &  {\bf 47.86} &  {\bf 56.42}  & 60.44   \\
%C3D & BSN+SNMS	& {\bf 29.58} & {\bf 37.38} &  {\bf 45.55} & {\bf 54.67} & 59.48   \\
\hline 
\hline
2Stream & TAG 	& 18.55 & 29.00  &  39.61 & - & -  \\
Flow & TURN		& 21.86 & 31.89 & 43.02 & 57.63  & 64.17  \\
2Stream & CTAP  & 32.49 & 42.61 & 51.97 & -  & -  \\
2Stream & BSN+NMS & 	 35.41 & 43.55 &  52.23 &  61.35 & 65.10  \\
2Stream & BSN+SNMS & 	 37.46 & 46.06 &  53.21 & 60.64 & 64.52 \\
\hline
2Stream & BMN+NMS & 	37.15 & 46.75 &  {\bf 54.84} & {\bf 62.19} & 65.22 \\
2Stream & BMN+SNMS & {\bf 39.36} & {\bf 47.72} &  54.70 &  62.07 & {\bf 65.49}  \\
%2Stream & BSN+SNMS & {\bf 37.46} & {\bf 46.06} &  {\bf 53.21} & 60.64 &  64.52   \\
\bottomrule
\end{tabular}
\label{table:proposal_thumos}
\vspace{-0.1cm}
\end{table}

The goal of proposal generation task is to generate high quality proposals to cover  action instances with high recall and high temporal overlap. 
To evaluate proposal quality, Average Recall (AR) under multiple IoU thresholds are calculated. Following   conventions,  IoU thresholds $[0.5:0.05:0.95]$ and $[0.5:0.05:1.0]$ are used for ActivityNet-1.3 and  THUMOS-14 separately. 
We calculate AR under different Average Number of proposals (AN) as AR@AN, and calculate the Area under the AR vs. AN curve (AUC) as metrics on ActivityNet-1.3, where AN is varied from 0 to 100.

\noindent
\textbf{Comparison with State-of-the-art Methods.}
%
%To verify the effectiveness of our method, we compare 
%
Table \ref{table:proposal_anet} demonstrates the proposal generation performance comparison on validation and testing set of ActivityNet-1.3, where our method significantly outperforms other proposal generation methods.
Especially, our method significantly improves AUC of validation set from $66.17\%$ to $67.10\%$ by $0.93\%$, which demonstrates that our method can achieve overall performance promotion.

\begin{table}[tbp]
\setlength{\belowcaptionskip}{-0.1cm} %缩小caption和下方文字的距离
\centering
\caption{  Ablation comparison between BSN \cite{lin2018bsn} and BMN in validation set of ActivityNet-1.3 in terms of AR@AN, AUC and inference speed. Inference speed here is the second (s) cost for processing a 3-minute video using a Nvidia 1080-Ti graphic card, including network inference time $T_{inf}$, proposal generation and proposal-feature generation (for BSN) time $T_{pro}$ and the total inference time $T_{sum} = T_{inf} + T_{pro}$.
\emph{e2e} here means modules of network are trained jointly.}
\small
\begin{tabular}{m{0.7cm}m{1.4cm}m{0.3cm}m{0.5cm}<{\centering}m{0.5cm}<{\centering}m{0.5cm}<{\centering}m{0.5cm}<{\centering}m{0.5cm}<{\centering}}
\toprule
Method & Module & \emph{e2e} & @100 & AUC & $T_{inf}$ & $T_{pro}$ & $T_{sum}$\\
\hline 
BSN & TEM     & -          & 73.57 & 64.80 & 0.002 & 0.034 & 0.036\\
BSN & TEM+PEM & $\times$   & 74.16 & 66.17 & 0.005 & 0.624 & {\bf 0.629}\\
\hline
BMN & TEM     & -          & 73.72 & 65.17 & 0.003 & 0.032 & 0.035\\
BMN & TEM+PEM & $\times$   & 74.36 & 66.43 & 0.007 & 0.062 & 0.069\\
BMN & TEM+PEM & \checkmark & 75.01 & 67.10 & 0.005 & 0.047 & {\bf 0.052}\\
\bottomrule
\end{tabular}
\label{table:ablation}
%\vspace{-0.45cm}
\end{table}

\begin{figure}[t]
\setlength{\abovecaptionskip}{-0.6cm} %缩小caption和图像之间的距离
\setlength{\belowcaptionskip}{-0.5cm} %缩小caption和下方文字的距离
\begin{center}
\begin{minipage}[b]{1.0\linewidth}
  \centering
  \centerline{\includegraphics[width=8.5cm]{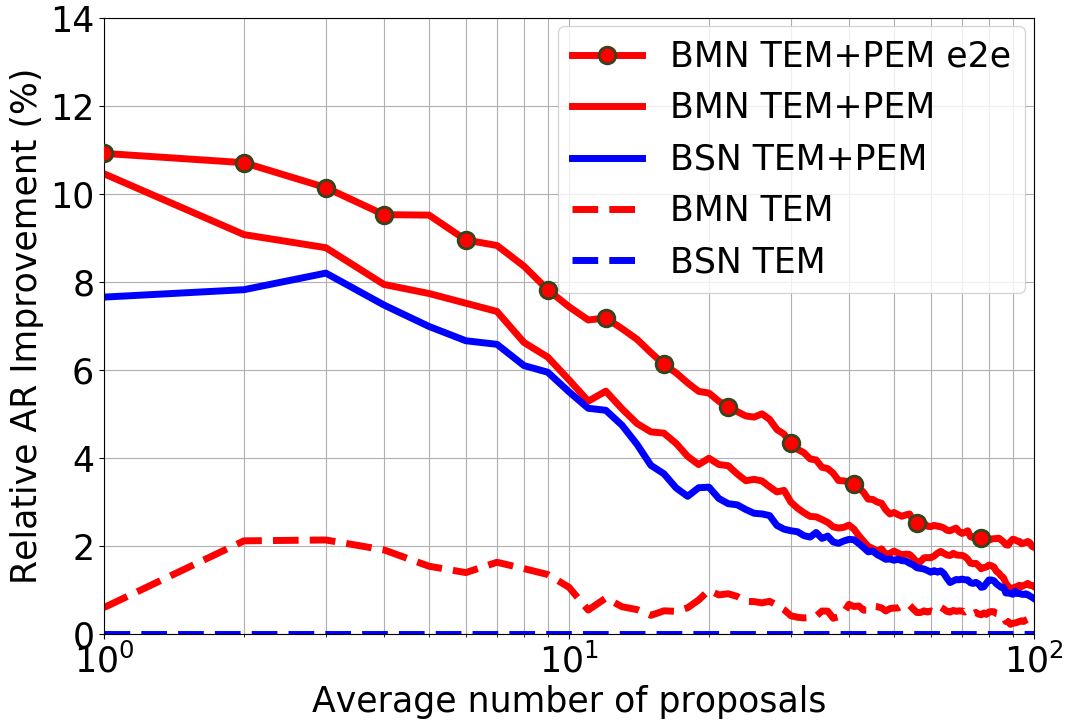}}
  \medskip
\end{minipage}
\end{center}
   \caption{Ablation comparison between BSN and BMN in terms of relative AR improvement ($\%$) vs AN curve on validation set of ActivityNet-1.3, where relative AR improvement is calculated based on BSN-TEM results.}
\label{fig:ablation}
\end{figure}

\begin{figure}[t]
\begin{center}
\begin{minipage}[b]{1.0\linewidth}
  \centering
  \centerline{\includegraphics[width=8.5cm]{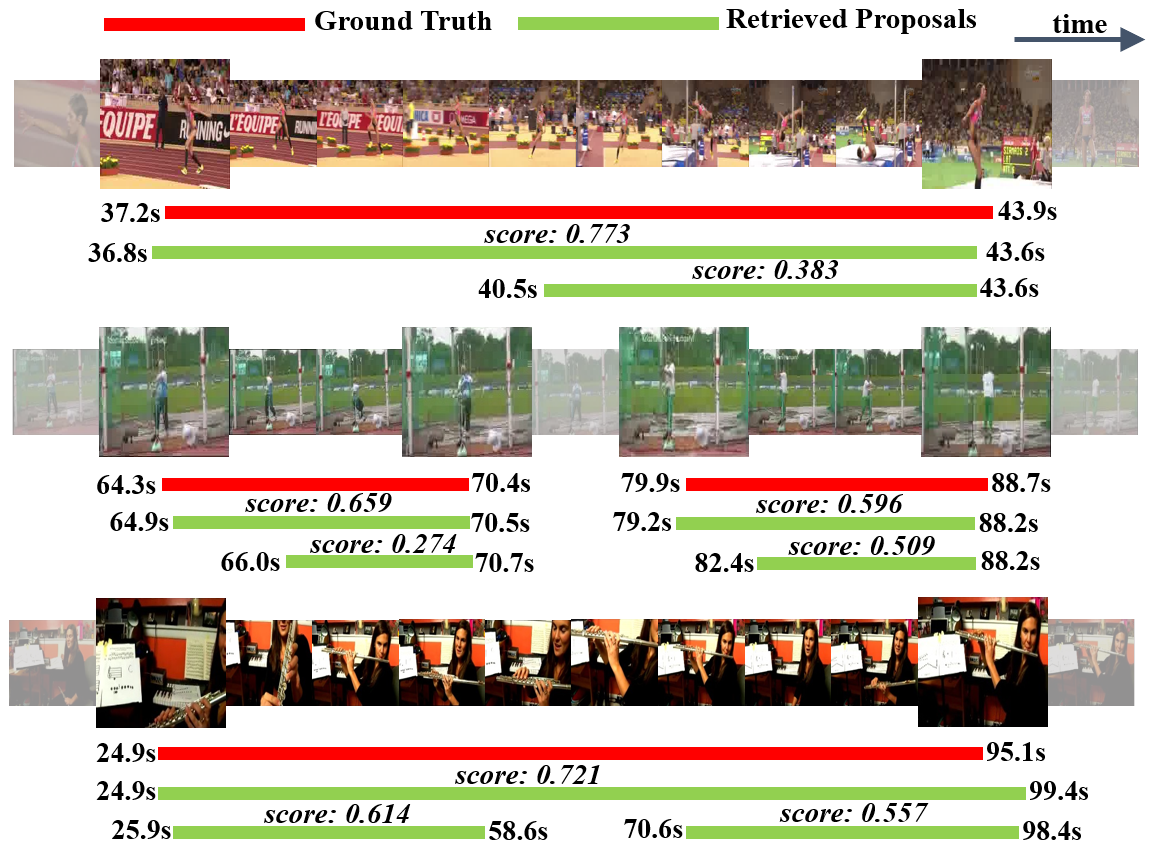}}
  \medskip
\end{minipage}
\end{center}
   \caption{Visualization examples of proposals and BM map generated by BMN on THUMOS-14 and ActivityNet-1.3 dataset.}
\label{fig:vis}
\vspace{-0.3cm}
\end{figure}

Table  \ref{table:proposal_thumos} demonstrates the proposal generation performance comparison on testing set of THUMOS-14. Since different feature encoding methods and redundant proposal suppression methods can affect performance largely, following BSN \cite{lin2018bsn}, we adopt both C3D and two-stream feature, both normal Greedy-NMS and Soft-NMS for fair comparison. 
Experiment results suggest that (1) based on either C3D or two-stream feature, our method outperforms other methods significantly when proposal number varies from 10 to 1000;
(2) no matter Greedy-NMS or Soft-NMS is adopted, our method outperforms other methods significantly; (3) Soft-NMS can improve  average recall performance especially under small proposal number, which is helpful for temporal action proposal generation task.
These results together suggest the effectiveness of our method and its effectiveness mainly due to its own architecture. Qualitative results are shown in Fig \ref{fig:vis}.

\noindent
\textbf{Ablation Comparison with BSN.}
To confirm the effect of the BM mechanism, we conduct more detailed ablation study and comparison of effectiveness and efficiency between BSN \cite{lin2018bsn} and BMN. To achieve this, we evaluate the proposal quality and speed of BSN and BMN under multiple ablation configuration.
The experiment results are shown in Table \ref{table:ablation} and Fig \ref{fig:ablation}, which demonstrate that:

\begin{enumerate}
\setlength{\itemsep}{0pt}
\setlength{\parsep}{0pt}
\setlength{\parskip}{0pt}
\item Under similar network architecture and training objective, TEMs of BSN and BMN achieve similar proposal quality  and inference speed, which provides a reliable comparison baseline;
\item Adding separately trained PEM, both BSN and BMN obtain significant performance promotion,  suggesting that PEM plays an important role in the ``local to global'' proposal generation framework;
\item Jointly trained BMN achieves higher recall and faster speed than separately trained BMN, suggesting the effectiveness and efficiency of overall optimization;
\item Adding separately trained PEM, BMN achieves significant faster speed than BSN, since BM mechanism can directly generate confidence scores for all proposals simultaneously, rather than one-by-one respectively in BSN. Thus, PEM based on BM mechanism is  more efficient than original PEM. Combining TEM and PEM jointly can further improve the efficiency.
\end{enumerate}

Thus, these ablation comparison experiments suggest the effectiveness and efficiency of our proposed Boundary-Matching mechanism and unified BMN network, which can generate reliable confidence scores for all proposals simultaneously in fast speed.

\begin{table}[tbp]
\setlength{\belowcaptionskip}{-0.5cm} %缩小caption和图像之间的距离
\centering
\caption{  Generalizability evaluation of BMN on validation set of ActivityNet-1.3 in terms of AR@AN and AUC.  }
\small
\begin{tabular}{m{2.0cm}m{1.0cm}<{\centering}m{1.0cm}<{\centering}m{1.0cm}<{\centering}m{1.0cm}<{\centering}}
\toprule
 & \multicolumn{2}{c}{Seen } & \multicolumn{2}{c}{Unseen }  \\
\hline 
Training Data &  AR@100  & AUC & AR@100 & AUC \\
\hline 
Seen+Unseen 	& 72.96 & 65.02 & 72.68 & 65.05 \\
Seen  		    & 72.47 & 64.37 & {\bf 72.46} & {\bf 64.47} \\
\bottomrule
\end{tabular}
\label{table:gene}
%\vspace{-0.45cm}
\end{table}

\noindent
\textbf{Generalizability of Proposals.}
As a proposal generation method, an important property is the ability of generating high quality proposals for unseen action categories.
To evaluate this property, following BSN \cite{lin2018bsn}, two un-overlapped action subsets: ``Sports, Exercise, and Recreation" and ``Socializing, Relaxing, and Leisure" of ActivityNet-1.3 are chosen as \emph{seen} and \emph{unseen} subsets separately. There are 87 and 38 action categories, 4455 and 1903 training videos, 2198 and 896 validation videos on \emph{seen} and \emph{unseen} subsets separately.
And we adopt C3D network \cite{tran2017convnet} pre-trained on Sports-1M dataset \cite{sports1m} for feature extraction, to guarantee the validity of experiments.
We train BMN with \emph{seen} and \emph{seen}+\emph{unseen} training videos separately, and evaluate both BMN models on \emph{seen} and \emph{unseen} validation videos separately.
Results in Table \ref{table:gene} demonstrate that the performance drop is very slight in unseen categories, suggesting that BMN achieves great generalizability to generate high quality proposals for unseen actions, and can learn a general concept of when an action may occur.

\subsection{Action Detection with Our Proposals}

Another important aspect of evaluating the proposal quality is to put proposals in temporal action detection framework and evaluate its detection performance.
Mean Average Precision (mAP) is adopted as the evaluation metric of temporal action detection task, where we calculate Average Precision (AP) on each action category respectively. mAP with IoU thresholds $\left \{0.5,0.75,0.95\right \}$ and average mAP with IoU thresholds $[0.5:0.05:0.95]$ are used on ActivityNet-1.3, while mAP with IoU thresholds $\left \{0.3,0.4,0.5,0.6,0.7 \right \}$ are used on THUMOS-14.

\begin{table}[tbp]
\centering
\caption{Action detection results on validation and testing set of ActivityNet-1.3, where our proposals are combined with video-level classification results generated by \cite{zhao2017cuhk}.  }
\small
\begin{tabular}{p{2.1cm}p{0.62cm}<{\centering}p{0.62cm}<{\centering}p{0.62cm}<{\centering}p{0.9cm}<{\centering}p{0.9cm}<{\centering}}
\toprule
 & \multicolumn{4}{c}{validation} & testing  \\
\hline
Method  & 0.5  &  0.75  & 0.95  & Average  & Average  \\
\hline
%Wang et al. \cite{wang2016uts}    & 42.28 & 3.76  & 0.05   & 14.85 & 14.62 \\
%SCC \cite{heilbron2017scc}   & 40.00 & 17.90  & 4.70   & 21.70 & 19.30 \\
CDC \cite{shou2017cdc}    & 43.83  & 25.88  & 0.21   & 22.77  & 22.90 \\
SSN \cite{xiong2017pursuit}    & 39.12 & 23.48  & 5.49  & 23.98 & 28.28 \\
Lin et al. \cite{lin2017temporal} & 44.39   & 29.65  & 7.09  & 29.17 & 32.26 \\
BSN \cite{lin2018bsn} + \cite{zhao2017cuhk} & 46.45  & 29.96 & 8.02  & 30.03 & 32.87 \\
\hline
Ours + \cite{zhao2017cuhk} & {\bf 50.07} & {\bf 34.78} & {\bf 8.29} & {\bf 33.85} & {\bf 36.42} \\
\bottomrule
\end{tabular}
\label{table_detection_anet}
%\normalsize
%\vspace{-0.3cm}
\end{table}

\begin{table}[tbp]
\centering
\caption{Action detection results on testing set of THUMOS14, where video-level classifier UntrimmedNet \cite{wang2017untrimmednets} and proposal-level classifier SCNN-Classifier \cite{shou2016action} are combined with proposals.}
\small
\begin{tabular}{p{1.4cm}p{1.4cm}p{0.52cm}<{\centering}p{0.52cm}<{\centering}p{0.52cm}<{\centering}p{0.52cm}<{\centering}p{0.52cm}<{\centering}}
\toprule
Method & classifier & 0.7 & 0.6 & 0.5 & 0.4 & 0.3  \\
\hline
SST \cite{sst_buch_cvpr17} & SCNN-cls 	& - & - & 23.0 & - &  -\\
TURN\cite{gao2017turn} & SCNN-cls 		& 7.7 & 14.6 & 25.6 & 33.2 &  44.1\\
BSN \cite{lin2018bsn} & SCNN-cls & 15.0 & 22.4 & 29.4 & 36.6 & 43.1 \\
Ours & SCNN-cls  & {\bf 17.0} & {\bf 24.5} & {\bf 32.2} & {\bf 40.2} & {\bf 45.7} \\
\hline
SST \cite{sst_buch_cvpr17} & UNet 		& 4.7 & 10.9 & 20.0 & 31.5 &  41.2\\
TURN\cite{gao2017turn} & UNet 			& 6.3 & 14.1 & 24.5 & 35.3 &  46.3\\
BSN \cite{lin2018bsn} & UNet & 20.0 & 28.4 & 36.9 & 45.0 & 53.5 \\
Ours & UNet & {\bf 20.5} & {\bf 29.7} & {\bf 38.8} & {\bf 47.4} & {\bf 56.0} \\ 
\bottomrule
\end{tabular}
\label{table_detection_thumos}
\normalsize
\vspace{-0.3cm}
\end{table}

To achieve this, we adopt the two-stage ``detection by classifying proposals'' temporal action detection framework to combine BMN proposals with state-of-the-art action classifiers.
Following BSN \cite{lin2018bsn}, on ActivityNet-1.3, we adopt top-1 video-level classification results generated by method \cite{zhao2017cuhk} and use confidence scores of BMN proposals for detection results retrieving. On THUMOS-14, we use both top-2 video-level classification results generated by UntrimmedNet \cite{wang2017untrimmednets}, and proposal-level SCNN-classifier to generate classification result for each proposal.
For ActivityNet-1.3 and THUMOS-14 datasets, we use first 100 and 200 temporal proposals per video separately.

The experiment results on ActivityNet-1.3 are shown in Table \ref{table_detection_anet}, which demonstrate that BMN proposals based detection framework significantly outperform other state-of-the-art temporal action detection methods.
The experiment results on THUMOS-14 are shown in Table \ref{table_detection_thumos}, which suggest that:
(1) no matter video-level or proposal-level action classifier is used, our method achieves better detection performance than other state-of-the-art proposal generation methods;
(2) using BMN proposals, video-level classifier \cite{wang2017untrimmednets} achieves significant better performance than proposal-level classifier \cite{shou2016action}, indicating that BMN can generate  confidence scores reliable enough for retrieving  results.

\section{Conclusion}

In this paper, we introduced the Boundary-Matching mechanism for evaluating confidence scores of densely distributed proposals, which is achieved via denoting proposal as BM pair and combining all proposals as BM confidence map. 
Meanwhile, we proposed the Boundary-Matching Network (BMN) for effective and efficient temporal action proposal generation, where BMN generates proposals with precise boundaries and flexible duration via combining high probability boundaries, and simultaneously generates reliable confidence scores for all proposals based on BM mechanism.
Extensive experiments demonstrate that BMN outperforms other state-of-the-art proposal generation methods in both  proposal generation and temporal action detection tasks, with remarkable efficiency and  generalizability.

{\small
\bibliographystyle{ieee}
\bibliography{egbib}
}

\end{document}